\documentclass[runningheads]{llncs}
\usepackage{hyperref}
\usepackage[T1]{fontenc}
\usepackage{graphicx}
\begin{document}

\title{Integrated Multimodal AI System for Retrieval-Augmented Reasoning, Object Sensing, and Damage Analysis}

\titlerunning{Integrated Multimodal AI System}

\author{Kalelo Dukuray\inst{1} \and
Israel Pina\inst{1} \and
Evan Perez\inst{1} \and Erika Ardiles-Cruz\inst{2}
\and Jie Wei\inst{1*}}
\authorrunning{Dukuray, Pina, Perez, Ardiles-Cruz, Wei}
\institute{ Dept. of Computer Science, City College of New York, New York, NY 10031, USA \and
Air Force Research Lab, Rome, NY 13441, USA\\
\email{jwei@ccny.cuny.edu} 
\footnote{\dag~This work is publicly released with No. AFRL-2026-3022.}\\
}
\maketitle              % typeset the header of the contribution
\begin{abstract}

This work presents a unified multimodal AI system for damage assessment that integrates retrieval-augmented generation (RAG) models, thermal spectrum perception, vision foundation model pipelines, and exploratory wireless signal sensing. 
A RAG component is developed to ground a locally hosted language model in project-specific documentation, including specialized damage level classification \cite{wei2021nida} criteria to mitigate hallucinations during inference. Controlled comparisons against static few-shot prompting demonstrate that dynamic retrieval improves grounding and factual consistency. We further compare vector-based RAG with a knowledge graph variant constructed via entity-relation extraction, and show that graph-based retrieval produces stronger responses for damage assessment queries requiring cross-document reasoning, motivating hybrid dense, sparse, and graph-aware retrieval.
To address limitations of EO imagery under adverse lighting and weather conditions, infrared (IR)/thermal sensing is employed for object detection and segmentation. Our detectors generate candidate detections, yielding improved segmentation of a broad array of objects. Paired IR versus visible spectrum tracking experiments reveal failure modes, motivating multimodal fusion for robust object detection and damage analysis. Vision foundation and vision-language models are leveraged to generate synthetic damage imagery and classify damage severity with high accuracy, supporting training and validation of downstream damage assessment models. Finally, exploratory Wireless-based sensing demonstrates potential to detect presence, motion, and post-event environmental changes where EO and IR sensing are ineffective. 
Overall, the system shows how RAG-grounded damage knowledge base, vision-based sensing, and wireless modalities complement one another to enable more robust, explainable, and accurate damage assessment across diverse scenarios, with promising results on public datasets.

\end{abstract}
\section{Introduction}
AI has been making breathtaking progress recently, affecting every aspect of our society. 
Pre‑trained language and vision models carry immense and highly useful knowledge that can dramatically amplify our work when applied carefully, yet the same power can mislead, bias, or derail results if used without rigorous oversight.
 In this work, we investigated the effective use of AI to integrate multimodal data, such as natural language, Electro-Optic (EO), InfraRed (IR), and wireless data, to achieve enhanced object sensing and damage analysis, which can find a wide array of applications in civilian and military applications. We explored a unified approach to integrate retrieval-augmented generation (RAG) models \cite{zhu2025knowledge} that can effectively mitigate hallucinations in the direct use of Large Language Models (LLMs) by inserting private or proprietary information or knowledge, e.g., private personnel information or classified documents that cannot be found in the pre-trained models, thermal spectrum perception using IR or near IR sensors to remedy EO signals under different lighting or situation condiations, the immense knowledge in the pre-trained Vision Foundation Models (VFMs) or Vision Languae Models (VLMs) are effectively exploited to help our knowledge fusion to achieve better object sensing and damage assessment. Under extreme situations, such as total destruction in natural disasters or battlefields, neither EO nor IR can yield useful detection results; we can then resort to exploratory wireless signal sensing to obtain useful object identification.

Three components fall within the scope of this research endeavor: 
1) 
We examined and evaluated two different Retrieval-Augmented Generation (RAG) approaches \cite{li2025investigating} in great detail, namely, a conventional vector store and a knowledge-graph-based variant built over the same corpus. Some hands-on results are shown to gain insights into the workings of these two principal RAG approaches. 
2)  The optimization of infrared (IR) data \cite{wei2013small} incorporated into the computer vision pipelines to enhance situational awareness in environments when Electro-Optical (EO) sensors are insufficient. Several different scenarios, such as human being detection, electric grid recognition, and Unmanned Aerial Vehicle (UAV) detection, are carefully studied to showcase the power of IR and EO fusion.
3)  We explored the use of wireless data-based sensing that demonstrates potential to detect presence, motion, and post-event environmental changes where EO and IR sensing are ineffective \cite{yuan2016novel}.

\begin{figure}
\centering
\includegraphics[width=0.8\textwidth]{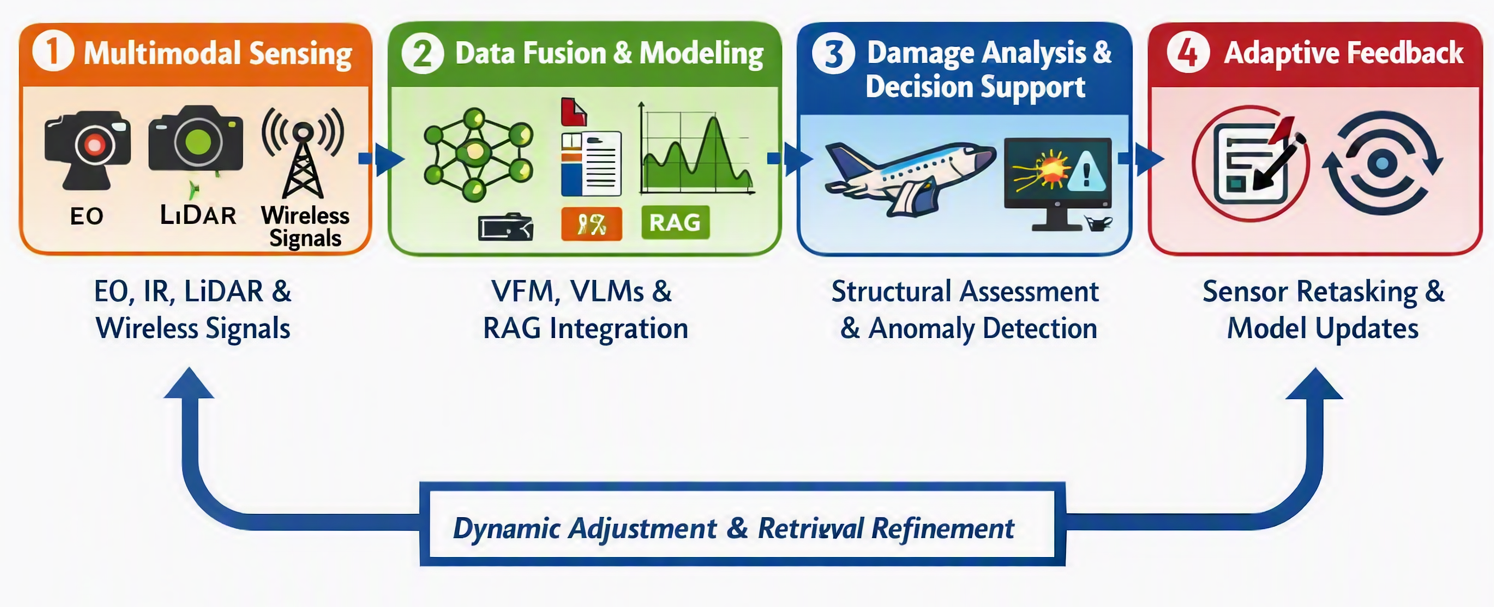}
\caption{DDDAS integration of this work: sensing, fusion, and damage analysis interact dynamically, enabling adaptive data acquisition and computational steering for resilient, real‑time decision support.} \label{fig0}
\end{figure}

As shown in Fig. \ref{fig0}, this integrated multimodal AI architecture naturally aligns with the Dynamic Data‑Driven Application Systems (DDDAS) paradigm, where sensing, modeling, and computation form a continuous adaptive loop. Each subsystem—RAG‑based reasoning, IR/EO vision fusion, and wireless sensing—embodies dynamic data‑driven feedback: real‑time sensor inputs refine model inference, and model outputs in turn guide subsequent data collection and fusion strategies. By coupling retrieval‑augmented reasoning with multimodal perception, the system exemplifies DDDAS principles of adaptive data acquisition and computational steering, enabling resilient damage assessment under evolving operational conditions.

 In the next three sections, the technical details, evaluation strategy, and observed results for each of the above three components are expanded at length, and algorithmic developments as well as empirical results are reported along the way. We conclude this paper with more remarks on this integrated multimodal AI system for effective retrieval-augmented reasoning, object sensing, and damage analysis.

\vspace*{-5mm}
\section{Vector and knowledge graph RAG}
To provide relevant information and knowledge and mitigate hallucinations by LLMs, which were pre-trained on general knowledge bases, the locally hosted RAG system is required to ground the query processing by LLMs. 

\noindent {\bf Methodology} In this work, two different RAG approaches are implemented: the vector store-based and knowledge-graph-based RAG, with their own pros and cons in grounding pre-trained LLMs. Our vector-based RAG: The backbone is llama3.2 served via Ollama, with nomic-embed-text supplying embeddings, Chroma serving as the vector store, and LangChain’s RecursiveCharacterTextSplitter chunking the corpus at a 250-token window with 20-token overlap. Conventional vector-based RAG works well for queries whose answer lives in a single chunk, but tends to underperform when the answer requires multi-hop ones that need to link entities that appear in different parts of the corpus, where the knowledge-graph-based method can work better. Knowledge-graph-based RAG: We used a NetworkX knowledge graph populated by entity-and-relation extraction over the same chunks as the knowledge-graph variant. Uploaded documents are ingested into two parallel stores in the same pass to construct the corresponding vector store and knowledge graph. At query time, the vector arm performs nearest-neighbor retrieval over Chroma in the usual way and pulls back the top-k chunks. Whereas the knowledge graph extracts entities from the user question, traverses the graph from those entities to gather connected nodes, and pulls back the chunks attached to them. Both contexts are then injected into the same model, and the two answers are streamed side by side. A per-query preference vote is persisted in a small SQLite table so that wins accumulate over time across collections, giving the comparison a lightweight empirical foundation rather than relying on anecdotal inspection.

\noindent{\bf Results} To compare the performance of these two different RAG techniques, forty test queries were issued against three ingested collections, namely, a mixture of project documentation, narrative test corpora, and short technical references,  and the per-query preference votes were recorded in each case. Queries were categorized post-hoc into two groups: 1) Single-passage queries, such as {\it ``What goes into Felix's harbor mocha, and what makes it the Saturday Secret?"},  whose answer is concentrated within one chunk of the source corpus, where Vector RAG can correctly locate the chunk with all details, whereas the graph RAG pulls chunks attached to the entity neighbors and likely to fragment passage-level information that misses the required details. 2) The multi-hop entity-linking queries, such as {\it ``How is Detective Castellan related to the founder of the Lantern Bookshop?"}, whose answer requires combining facts from two or more chunks via a shared entity, since the correct answer cannot be found directly in a single chunk but must be assembled from at least 3 different chunks, which can only be achieved by the reasoning power of the knowledge graph. The aggregated outcomes are summarized in Table~\ref{tab_rag}. The results match the architectural prediction. On single-passage queries, where nearest-neighbor retrieval can land the entire answer in one chunk, the vector arm wins, or ties on roughly four-fifths of the tested queries, and the knowledge-graph machinery adds little. On multi-hop entity-linking queries, where the answer is distributed across the corpus and requires combining facts via a shared entity, the knowledge-graph arm wins on more than 80\% of queries, and the vector arm wins on only two. Overall, across both query types, the knowledge-graph variant wins approximately twice as often as the vector variant, with the rest split between ties and vector wins.

\begin{table}
\centering
\caption{Per-query preference votes across 40 test queries, categorized by whether the answer required combining facts from multiple chunks}\label{tab_rag}
\begin{tabular}{|l||c|c|c|c|}
\hline
{\bf Query Type} &  {\bf Total Queries} & {\bf Vector RAG Wins} & {\bf Graph RAG Wins} & {\bf Ties}\\
\hline \hline
{\bf Single-passage} & 22 & 9  & 4 & 9 \\ \hline
{\bf Multi-hop} & 18 & 2 & 15 & 1 \\ \hline
{\bf Overall} & 40 & 11 & 19 & 10 \\ \hline
\end{tabular}
\end{table}

Across the votes gathered so far, the knowledge-graph variant has produced noticeably stronger generations on questions whose answers depend on linking entities that appear in different parts of the corpus — exactly the regime in which flat cosine similarity over chunk embeddings tends to miss the relevant material. On questions whose answers are concentrated within a single passage, the two arms remain roughly comparable, and the simpler vector pipeline is a reasonable default. The setup also points to clear directions for further improvement. On the retrieval side, hybrid sparse-dense search and graph-aware re-ranking are both natural extensions of the same architecture. On the generation side, swapping the local model for an instruction-tuned model trained with reinforcement learning from human feedback (RLHF) would let the same retrieved context be exploited by a generator that is in itself better aligned with human preferences. 

Our RAG system is also {\it adaptable}: Users can input and correct the answers of the RAG system so that the corresponding vector and graph RAG can enrich their knowledge base with human feedback in the loop. From our rigorous tests, this adaptive vector and graph RAG system can greatly enhance the usefulness of the LLM in natural language-based reasoning and knowledge grounding.

\section{IR modality for detection and segmentation}
\subsection{IR-based object detection}
In this line of work, we would like to establish a baseline of using cutting-edge VLMs and VFMs to classify and segment objects of interest, such as human beings at different scales and electric grid components, using IR and EO data.

\noindent{\bf Methodology} Using state-of-the-art YOLOv8 architectures \cite{varghese2024yolov8}, the Segment Anything Model (SAM) \cite{kirillov2023segment}, and multimodal fusion techniques, this work establishes a robust framework for autonomous detection and high-fidelity segmentation. The investigation utilized three distinct YOLOv8-based models, each optimized for unique operational domains:
1)
	HIT-UAV (High-altitude Infrared Thermal UAV): Designed for aerial thermography, this model facilitates search-and-rescue and nocturnal surveillance by detecting small-scale objects (persons, vehicles) from significant altitudes. 
2) FLIR-ADAS (Advanced Driver Assistance Systems): Focused on ground-level navigation, this model identifies pedestrians and obstacles in adverse environmental conditions where traditional RGB, LiDAR, or radar may experience signal degradation. 
3)	Electrical Grid Proliferation: A specialized industrial model trained to identify critical infrastructure components (e.g., insulators, breakers, and current transformers) to facilitate autonomous maintenance in low-visibility settings. 
The selection of the Vision Foundation Model (VFM): YOLOv8 (You Only Look Once) was predicated on its ability to utilize a single convolutional neural network for simultaneous bounding box prediction and class probability estimation, ensuring the real-time temporal resolution required for military and industrial assessment. 

\begin{figure}
\begin{center}
\includegraphics[width=0.7\textwidth]{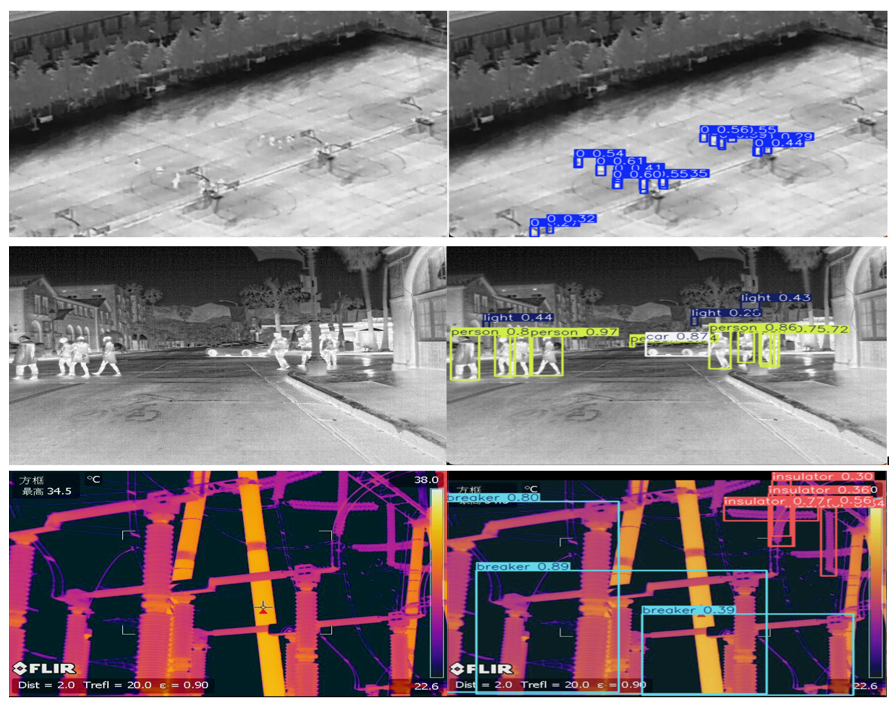}
\end{center}
\caption{IR-based object detection results for HIT-UAV (Row 1), FLIR-ADAS (Row 2), and Electrical Grid Proliferation (Row 3) datasets:  Each image on the right shows the result of running each of the three datasets' dedicated trained model on the test image illustrated on the left.} \label{fig2}
\end{figure}

\noindent{\bf Results} To move beyond simple bounding box localization, a Two-Stage Segmentation Pipeline \cite{wei2001image} was implemented using one Vision Language Model (VLM)~\cite{wei2025effective}: the Segment Anything Model (SAM). 1) Prompt-Guided Masking: YOLO-generated bounding boxes serve as spatial prompts for SAM, enabling class-specific segmentation without manual intervention. 2) Morphological Refinement: The pipeline utilizes automated opening and closing operations to mitigate noise and fill internal discontinuities, resulting in "production-ready" masks with object masks anchored by EO-based YOLO. 3) Intelligent Asset Harvesting: The system calculates object centroids and generates RGBA transparent crops, facilitating the automated creation of augmented datasets for downstream research. In Fig.~\ref{fig2}, the IR-based object detection and segmentation results are illustrated.

\subsection{Paired IR vs EO UAV tracking}
Reliable UAV tracking has applications across surveillance and air defense, but it is rarely robust to environmental conditions when carried by a single sensor. EO sensors struggle at night, in smoke or fog, against glare, and at long standoff. IR sensing covers many of those failure modes, but not all of them. Quantifying where each modality wins and which loses on the same input is necessary input for designing a multi-modal tracker rather than continuing to lean on either sensor alone.

\noindent{\bf Methodology} A paired-modality evaluation harness runs two YOLO detectors against the same aerial sequence, frame by frame: one trained on infrared imagery, one trained on visible-spectrum imagery. Both run at matched confidence and IoU thresholds so that any divergence in the results reflects modality and not hyperparameter choice. The program was applied to forty sequences from the Anti-UAV dataset: a single-class UAV detection benchmark with synchronized IR/visible pairs spanning daytime, low-light, and nighttime recording conditions. For each selected sequence, the two annotated streams are rendered side by side, accompanied by per-frame detection counts, detection rates, average and maximum confidence, and the inter-modality delta. A sequence was attributed to a winning modality only when its detection rate exceeded the other's by more than five percentage points; otherwise, the sequence was classified as a tie. The parameters of the harness are summarized in Table~\ref{tab_uav}, where it can be observed that although IR achieved better results than EO in far more sequences, fusing EO with IR can essentially deliver better results with a mean detection rate boosted from 91.8\% to 94.2\%.

\begin{table}
\centering
\caption{Aggregate detection performance across 40 Anti-UAV sequences. Fused = per-frame union of IR-based and EO-based detections.}\label{tab_uav}
\begin{tabular}{|l||c|c|c|}
\hline
{\bf Metrics} &  {\bf IR-only} & {\bf EO-only} & {\bf IR and EO Fused}\\
\hline \hline
{\bf Mean detection rate \%} & 91.8 & 44.2  & 94.2 \\ \hline
{\bf Sequences Wins} & 24 & 4 &  \\ \hline
{\bf Sequences Ties} & 12 & 12 & \\ \hline

\end{tabular}
\end{table}

\noindent{\bf Results} A consistent pattern emerges across sequences.  Neither modality is sufficient on its own: EO fails catastrophically at night and in low-light conditions, and completely fails to detect in eleven of forty sequences, while IR fails when the object sits against a thermally similar background, such as sun-warmed urban architecture during the middle of the day. The two failure modes are largely disjoint, which is exactly what makes simple union-level fusion effective: the 94.2\% fused detection rate exceeds either single-modality stream and recovers the EO-only frames that an IR-only tracker would miss. The data motivates a multi-modal tracking stack rather than a single-sensor variant as the operational design, and points naturally toward incorporating further modalities, e.g., audio~\cite{jiao2023audio}, as the next step toward UAV tracking that remains reliable under varied environmental and clutter conditions.

\section{Exploratory Wireless Data Sensing}
In natural disasters and battlefield damage assessment, use cases will arise when none of the EO, IR, and LiDAR sensors is sufficient: when the buildings are totally damaged according to EO or IR, no meaningful objects can be detected/segmented, but the search for survivors in natural disasters and still active terrorists’ activities under the damages regions are still of crucial importance, this is the case when wireless signal detection can be a valuable additional data modality that is ready to help. 

\noindent{\bf Methodology} The wireless signals, unlike EO and IR sensor data, are not impacted unless the emitters are physically damaged. The presence and/or absence of different emitters paint the complete picture of the wireless signatures in the impacted regions, thus providing the presence, motion, and post-event environmental changes where EO, IR, and LiDAR sensing \cite{thompson2023multi} are of no use. As the strengths of different wireless signals are distance-dependent, the active movements of this system can effectively locate the locations of the emitters by triangulation methods.

\begin{figure}
\centering
\includegraphics[width=0.65\textwidth]{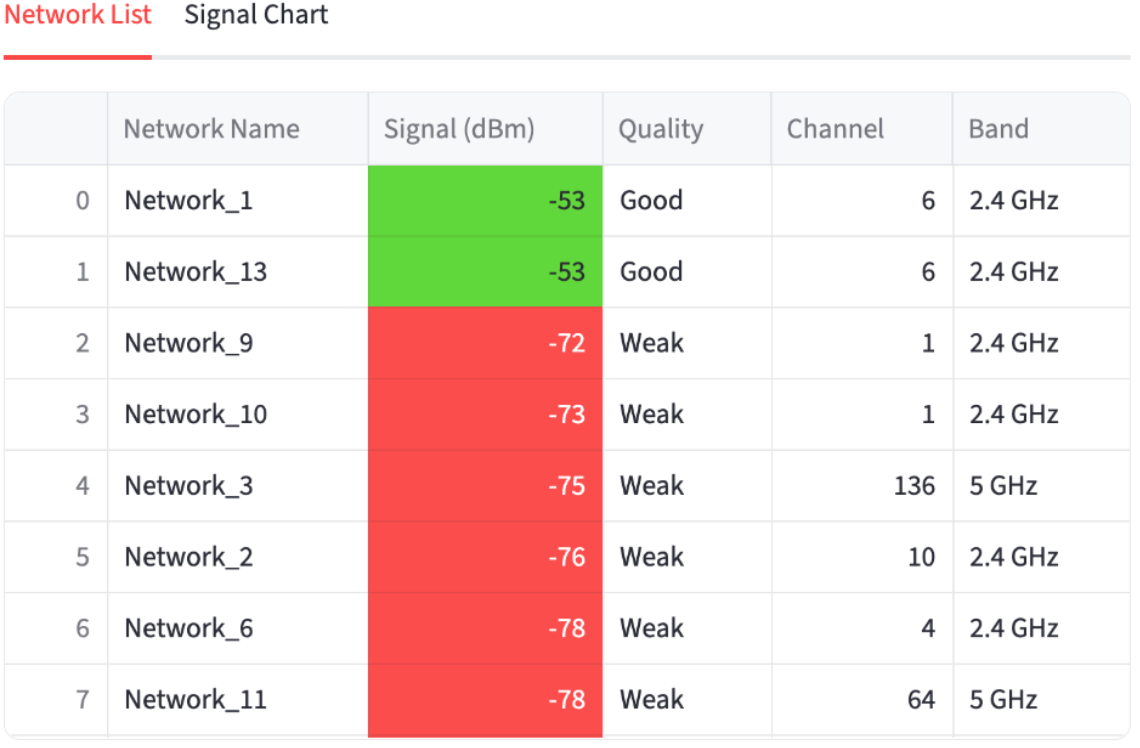}
\caption{The detected wireless emitters' frequencies and strengths are a valuable detector when other data modalities are of no power in case of total damage.} \label{fig4}
\end{figure}

\noindent{\bf Results} In Fig.\ref{fig4}, the wireless signal strengths of different wireless emitters are reliably detected, which is not affected by the damage levels of buildings in any way. 

\section{Conclusion} 
In this work, using cutting-edge AI approaches and models, we developed a system that can exploit the advantages of different data modalities, such as natural language, EO, IR, and wireless signals, to deliver better object detection or damage analysis in civilian and military applications.
In all the investigated use cases, a single-representation baseline is replaced or augmented with a structured combination of complementary signals:  IR plus EO streams in the object detection and UAV tracking work, vector retrieval plus a knowledge graph in the RAG work, and wireless signature detections when all other data modalities are powerless in the case of total damage. In each case, the combined system produces measurable gains where the simpler singular data modality cannot deliver. 
Taken together, these contributions reinforce the broader argument that practical performance on real sensing and reasoning tasks tends to come less from any single dominant model and more from arranging complementary signals so that the failure modes of one are covered by the strengths of another. In this work, the use of LLM, vector, and graph RAG, VFM, and VLM is demonstrated with real use cases as concrete evidence that cutting-edge AI approaches, if used properly, can significantly enhance our tasks.

% \begin{table}
% \caption{Table captions should be placed above the
% tables.}\label{tab1}
% \begin{tabular}{|l|l|l|}
% \hline
% Heading level &  Example & Font size and style\\
% \hline
% Title (centered) &  {\Large\bfseries Lecture Notes} & 14 point, bold\\
% 1st-level heading &  {\large\bfseries 1 Introduction} & 12 point, bold\\
% 2nd-level heading & {\bfseries 2.1 Printing Area} & 10 point, bold\\
% 3rd-level heading & {\bfseries Run-in Heading in Bold.} Text follows & 10 point, bold\\
% 4th-level heading & {\itshape Lowest Level Heading.} Text follows & 10 point, italic\\
% \hline
% \end{tabular}
% \end{table}

% \noindent Displayed equations are centered and set on a 
% line.
% \begin{equation}
% x + y = z
% \end{equation}
% Please try to avoid rasterized images for line-art diagrams and
% schemas. Whenever possible, use vector graphics instead (see
% Fig.~\ref{fig1}).

% \begin{figure}
% \includegraphics[width=\textwidth]{fig1.eps}
% \caption{A figure caption is always placed below the illustration.
% Please note that short captions are centered, while long ones are
% justified by the macro package automatically.} \label{fig1}
% \end{figure}

\subsubsection{Acknowledgements} The authors would like to gratefully acknowledge the support of ML-RCP and AFRL FA950-21-1-0082.

\bibliographystyle{splncs04}
\bibliography{mybib}




% LNCS Homepage, \url{http://www.springer.com/lncs}. Last accessed 4
% \end{thebibliography}
\end{document}